\documentclass[sn-mathphys,Numbered, iicol]{sn-jnl}% Math and Physical Sciences Reference Style
\usepackage{graphicx}%
\usepackage{multirow}%
\usepackage{amsmath,amssymb,amsfonts}%
\usepackage{amsthm}%
\usepackage{mathrsfs}%
\usepackage[title]{appendix}%
\usepackage{xcolor}%
\usepackage{textcomp}%
\usepackage{manyfoot}%
\usepackage{booktabs}%
\usepackage{algorithm}%
\usepackage{algorithmicx}%
\usepackage{algpseudocode}%
\usepackage{listings}%

\RequirePackage{xspace}
\usepackage{graphicx}
\usepackage{xcolor}
\usepackage{amsmath}
\usepackage{amssymb}
\usepackage{multirow}
\usepackage{hyperref}

\newcommand*\tcircle[1]{%
  \raisebox{-0.5pt}{%
    \textcircled{\fontsize{7pt}{0}\fontfamily{phv}\selectfont #1}%
  }%
}

\makeatletter
\DeclareRobustCommand\onedot{\futurelet\@let@token\@onedot}
\def\@onedot{\ifx\@let@token.\else.\null\fi\xspace}

\def\etal{\emph{et al}\onedot}

\theoremstyle{thmstyleone}%
\theoremstyle{thmstyletwo}%

\theoremstyle{thmstylethree}%

\begin{document}

\title[Article Title]{ViStripformer: A Token-Efficient Transformer for Versatile Video Restoration}


\author[1]{\fnm{Fu-Jen} \sur{Tsai}}\email{fjtsai@gapp.nthu.edu.tw}

\author[2]{\fnm{Yan-Tsung} \sur{Peng}}\email{ytpeng@cs.nccu.edu.tw}

\author[3]{\fnm{Chen-Yu} \sur{Chang}}\email{david.cs08@nycu.edu.tw}

\author[3]{\fnm{Chan-Yu} \sur{Li}}\email{frank.cs08@nycu.edu.tw}

\author[4]{\fnm{Chung-Chi} \sur{Tsai}}\email{chuntsai@qti.qualcomm.com}

\author[3]{\fnm{Yen-Yu} \sur{Lin}}\email{lin@cs.nycu.edu.tw}

\author*[1]{\fnm{Chia-Wen} \sur{Lin}}\email{cwlin@ee.nthu.edu.tw}

\affil[1]{\orgdiv{Department
of Electrical Engineering}, \orgname{National Tsing Hua University}, \orgaddress{\city{Hsinchu}, \postcode{300048},  \country{Taiwan}}}

\affil[2]{\orgdiv{Department of Computer Science}, \orgname{National Chengchi University}, \orgaddress{\city{Taipei}, \postcode{116011}, \country{Taiwan}}}

\affil[3]{\orgdiv{Department of Computer Science}, \orgname{National Yang Ming Chiao Tung University}, \orgaddress{\city{Hsinchu}, \postcode{300093}, \country{Taiwan}}}

\affil[4]{\orgdiv{Qualcomm Technologies}, \orgname{Inc}, \orgaddress{\city{San Diego, CA}, \postcode{92121}, \country{USA}}}


\abstract{Video restoration is a low-level vision task that seeks to restore clean, sharp videos from quality-degraded frames. One would use the temporal information from adjacent frames to make video restoration successful.
Recently, the success of the Transformer has raised awareness in the computer-vision community. However, its self-attention mechanism requires much memory, which is unsuitable for high-resolution vision tasks like video restoration. 
In this paper, we propose ViStripformer (Video Stripformer), which utilizes spatio-temporal strip attention to catch long-range data correlations, consisting of intra-frame strip attention (Intra-SA) and inter-frame strip attention (Inter-SA) for extracting spatial and temporal information.
It decomposes video frames into strip-shaped features in horizontal and vertical directions for Intra-SA and Inter-SA to address degradation patterns with various orientations and magnitudes.
Besides, ViStripformer is an effective and efficient transformer architecture with much lower memory usage than the vanilla transformer.  
Extensive experiments show that the proposed model achieves superior results with fast inference time on video restoration tasks, including video deblurring, demoiréing, and deraining.}




\maketitle

\section{Introduction}

Many factors could cause videos to have low quality, such as camera shakes, moving or rainy scenes, and electronic interferences, which would make viewers unable to see the recorded scenes clearly.  There has been many research works proposed to restore such videos~\cite{Pan_2020_CVPR, zhong2022real, zhang2022spatio, wang2022MMP, Suin_2021_CVPR, Isobe_2020_CVPR, dai2022video, Li_2021_CVPR, MANA, Huang_2022_CVPR, Wang_2019_CVPR_Workshops, chao2022, Yang_2019_CVPR_Dual_Flow, 9439949, Liu_2018_CVPR, su2017deep, frvsr}. 
Compared to single image restoration tasks~\cite{Zamir2021MPRNet, Wang_2022_CVPR, Zamir2021Restormer, MIMO, AirNet, IPT, Tsai2022Stripformer, yang2020learning, Chi_2021_CVPR, SAPN2020}, video restoration can utilize not only spatial but temporal information from videos to achieve better results. In recent years, most video restoration approaches~\cite{Pan_2020_CVPR, zhang2022spatio, chao2022, Jiang_2022_ECCV, zhong2022real, 9439949, Huang_2022_CVPR, patil2022video, dai2022video} were developed based on Convolutional Neural Networks (CNNs). However, CNN-based models may not be able to extract long-range features easily to fully utilize temporal information for video restoration, thus only achieving sub-optimal performance.

Also, objects and scenes are often misaligned in consecutive frames for degraded videos, requiring restoration methods to process misaligned adjacent frames. Some methods resort to image warping based on optical flows to align objects and scenes for better utilizing temporal information extracted from consecutive frames~\cite{Kim_2015_CVPR, su2017deep, Li_2021_CVPR, Pan_2020_CVPR, Huang_2022_CVPR, frvsr}. However, in regions with large displacements, motion blurs, noise, etc., optical flows could be inaccurately estimated, leading to unsatisfactory deblurring results. Besides, estimating optical flows requires additional computations and causes longer latency. Some methods utilize a recurrent neural network (RNN) to propagate and process temporal features~\cite{chao2022}, but it often leads to undesirable error accumulation~\cite{zhou2021rta}.

\begin{figure}
\begin{center}
\includegraphics[width=0.9\columnwidth]{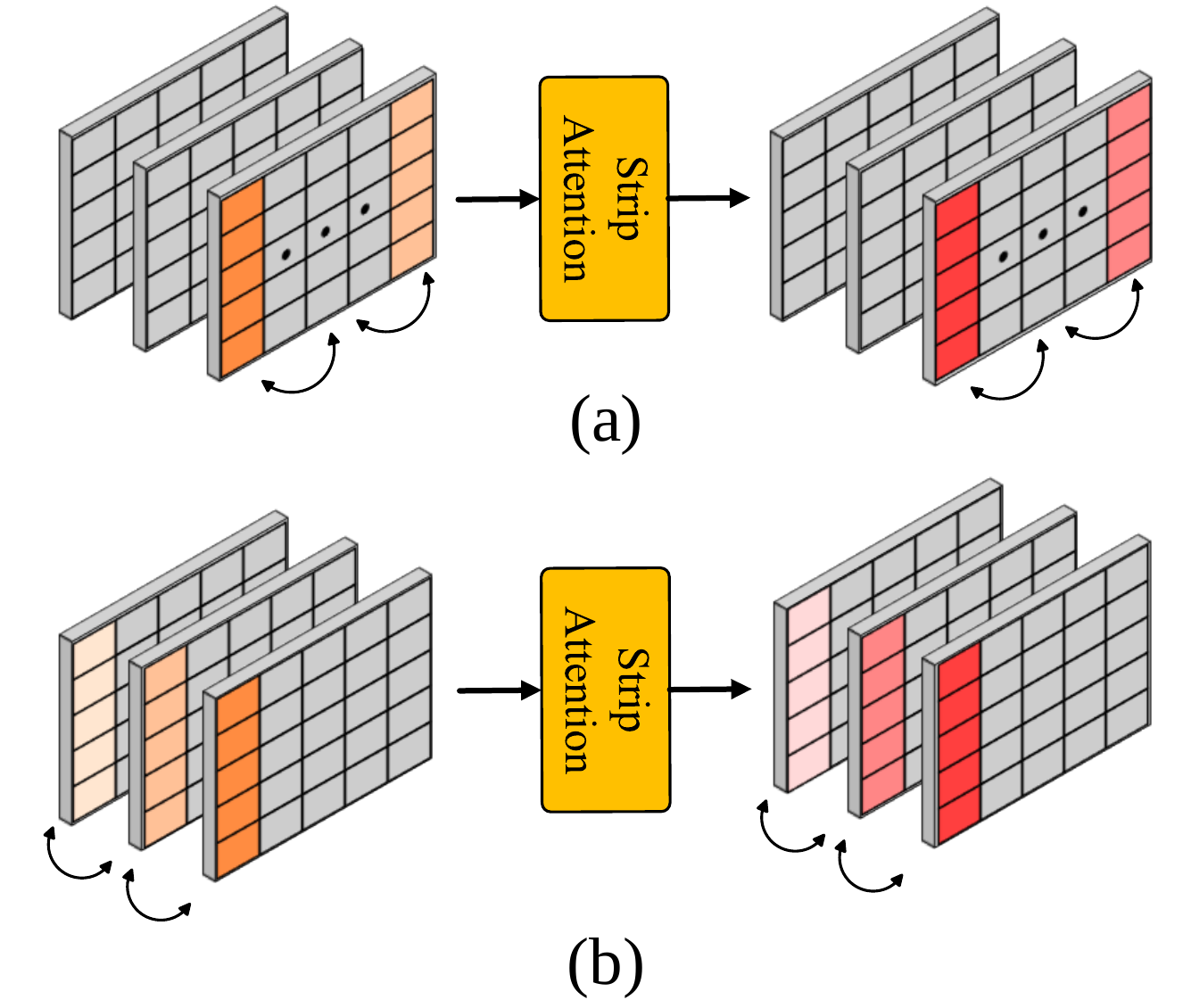}
\end{center}
%\vspace{-0.1in}
\caption{Concept of strip attention. (a) Vertical Intra-frame Strip Attention (V-Intra-SA) captures spatial information in a frame. (b) Vertical inter-frame strip attention (V-Inter-SA) extracts temporal information from the corresponding strips across consecutive frames. The horizontal intra-frame and inter-frame strip attentions are designed symmetrically.  
}
\label{fig:teaser}
\end{figure}

Recent research attempts on computer vision tasks using Vision Transformers~\cite{vaswani2017attention} have elaborated great potential, including high-level vision~\cite{dosovitskiy2020vit, liu2021Swin, chu2021Twins, chen2022regionvit, DETR, Ranftl21, zhu2021deformable, xie2021segformer} and low-level vision~\cite{Tsai2022Stripformer, IPT, Zamir2021Restormer, Wang_2022_CVPR, fgst, yang2020learning, yan2020sttn} tasks. Transformers feature the self-attention mechanism and token mixer to exploit long-range dependencies, which can help correlate spatial and temporal features globally for video restoration tasks. Nevertheless, the burdensomely high memory usage of the self-attention mechanism~\cite{Wang_2018_CVPR}, up to $\mathcal{O}((HWT)^2)$, where $H$, $W$, and $T$ respectively denote the height, width, and the number of frames of an input video, limits its applicability to vision tasks on high-resolution videos.

\begin{figure}
\begin{center}
\includegraphics[width=1\columnwidth]{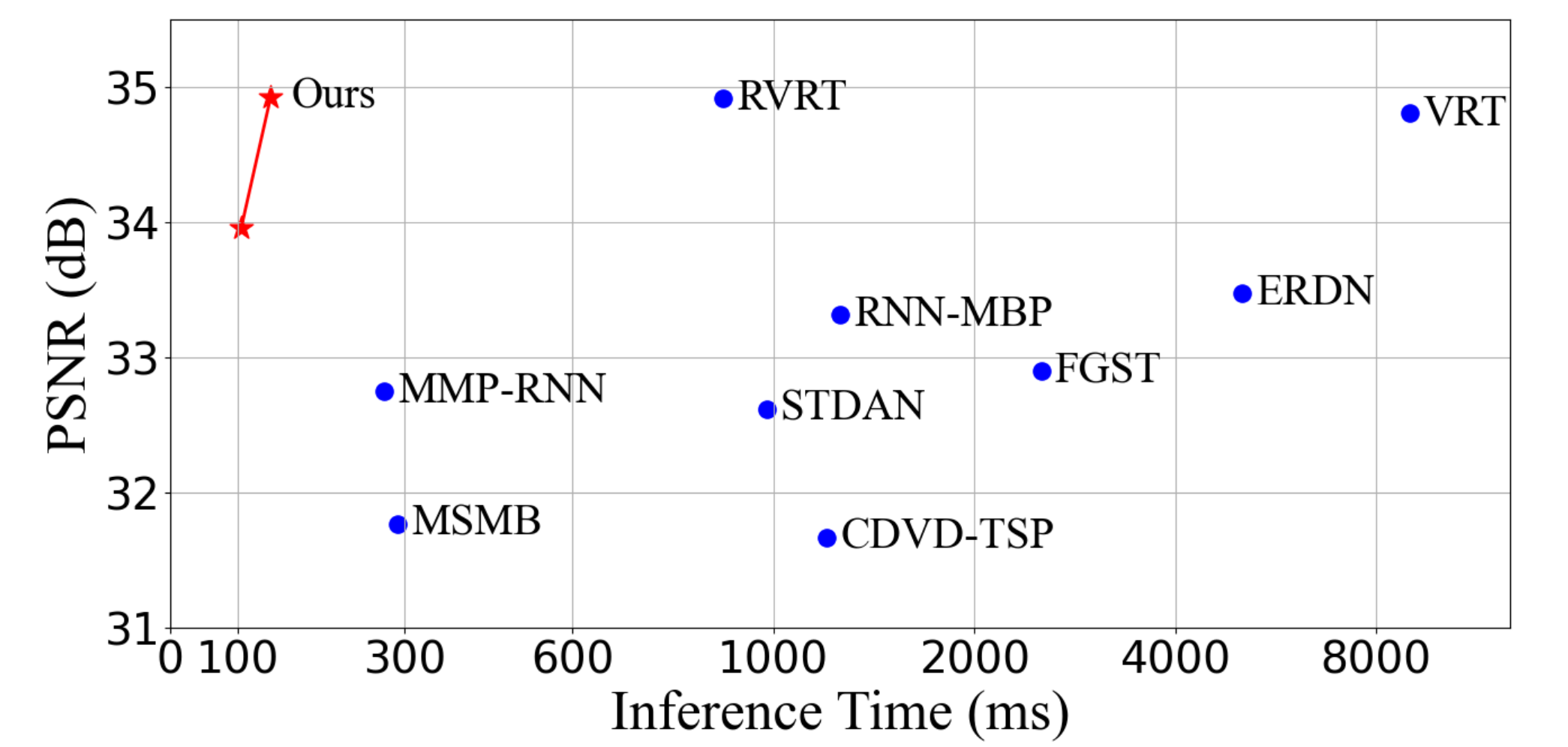}
\end{center}
%\vspace{-0.2in}
\caption{Quality-runtime performance comparison of representative methods on the GoPro test set. Our method runs effectively with low latency. 
}
\label{fig:PSNR_Curve}
\end{figure}

To address the above-mentioned high demand for memory, extended from our Stripformer~\cite{Tsai2022Stripformer},  we propose ViStripformer (Video Stripformer) with spatio-temporal strip attention (STSA)  blocks for video restoration. An STSA block consists of an Intra-frame Strip Attention (Intra-SA) block and an Inter-frame Strip Attention (Inter-SA) block to capture spatial and temporal information, as shown in Figs.~\ref{fig:teaser}(a) and~\ref{fig:teaser}(b), respectively.
The Intra-SA performs strip-wise attention within each frame, capturing spatial information. In contrast, the Inter-SA applies strip-wise attention to the co-located strips across multiple frames, extracting temporal information from consecutive video frames. The Intra-SA and Inter-SA only require memory usage of $\mathcal{O}((H^2+W^2)T)$ and $\mathcal{O}((H+W)T^2)$, significantly less than the vanilla transformer's $\mathcal{O}((HWT)^2)$. 
Furthermore, strip-wise attention catches degraded patterns with horizontal and vertical strip-shaped features like projection to deal with them in diverse orientations and magnitudes, making ViStripformer particularly suitable for restoring directional degradation patterns, such as blur, rain, and moiré. Thus, the proposed ViStripformer with interweaving multi-head Intra-SA and Inter-SA blocks works well for restoring various video restoration tasks. 
In addition, ViStripformer can handle misaligned frames without relying on optical flows or recurrent neural networks (RNNs), thus leading to a faster inference time and better performance than the previous video restoration methods, as shown in Fig.~\ref{fig:PSNR_Curve}.

Our main contributions can be summarized as follows: 
\begin{itemize}
\item We propose ViStripformer, utilizing intra-frame and inter-frame strip attentions to extract spatio-temporal features for video restoration, especially suitable for dealing with directional degradation patterns. .
\item The proposed Intra-SA and Inter-SA are with the memory usage of $\mathcal{O}((H^2+W^2)T)$ and $\mathcal{O}((H+W)T^2)$, respectively, much more efficient than the vanilla transformer's $\mathcal{O}((HWT)^2)$. 
\item Extensive experiments show that ViStripformer achieves superior performance with fast inference time for video deblurring, deraining, and demoiréing tasks.
\end{itemize}

ViStripformer is an extension of our previous work, Stripformer~\cite{Tsai2022Stripformer}, which was specifically designed for image deblurring. In this work, we conduct the following modifications to enhance the capabilities of strip attention: \textbf{(i)} We extend the strip attention to the intra-frame and inter-frame strip attentions to efficiently and effectively address spatio-temporal features. \textbf{(ii)} We apply ViStripformer to three video restoration tasks, including deblurring, deraining, and demoiréing, and the proposed framework achieves superior performances on all three tasks.

\section{Related Work}

\subsection{Video Restoration}
Recently, video restoration~\cite{Pan_2020_CVPR, zhong2022real, zhang2022spatio, wang2022MMP, Suin_2021_CVPR, Isobe_2020_CVPR, dai2022video, MANA, Huang_2022_CVPR, Wang_2019_CVPR_Workshops, chao2022, Yang_2019_CVPR_Dual_Flow, 9439949, Liu_2018_CVPR, frvsr} has been drawing much attention for its many practical applications. Compared to image restoration~\cite{Zamir2021MPRNet, Wang_2022_CVPR, Zamir2021Restormer, MIMO, AirNet, IPT, Tsai2022Stripformer, yang2020learning, Chi_2021_CVPR, SAPN2020}, video restoration can utilize both spatial and temporal information to facilitate video recovery. This paper proposes Vistripformer (Video Stripformer) to deal with directional degradation patterns with various orientations and magnitudes, such as blur, rain, and moiré. Next, we describe existing video restoration methods for video deblurring, deraining, and demoiréing.

\vspace{0.1in}
\noindent\textbf{Video Deblurring}
Most existing video deblurring methods use a set of consecutive frames to restore the center frame in that set~\cite{Pan_2020_CVPR, zhong2022real, zhang2022spatio, wang2022MMP, Suin_2021_CVPR, chao2022}. However, these frames may need to be aligned with the center frame for more accurate deblurring.
To align consecutive frames, one can apply image warping based on optical flows~\cite{Pan_2020_CVPR, liang2022vrt, liang2022rvrt}. Pan~\etal~\cite{Pan_2020_CVPR} adopted a pre-trained PWC Net~\cite{Sun2018PWC-Net} to estimate optical flows for warping adjacent frames to the target frame.  
However, estimating optical flows from quality-degraded frames may not be accurate and would fail when a scene change occurs~\cite{dai2022video}. Wang~\etal~\cite{wang2022MMP} proposed to predict bi-directional optical flows from high-frequency sharp frames and generate a motion magnitude prior for deblurring.
Also, RNN-based methods~\cite{zhong2022real, wang2022MMP, chao2022} have been commonly used in modeling spatio-temporal correlations, which can propagate features from the previous or future frames for alignment. 
Zhu~\etal~\cite{chao2022} devised a multi-scale bidirectional propagation module to fuse forward and hidden states, effectively exploiting unaligned neighboring frames.
Pan~\etal~\cite{Pan_2023_CVPR} proposed a  wavelet-based feature propagation that performs bi-directional feature propagation on low-frequency features.
Li~\etal~\cite{Li_2023_CVPR} proposed a grouped spatial-temporal shift operation that utilizes large-scaled spatial shifts and bi-directional temporal shifts to model temporal correspondences.   
Nevertheless, the bidirectional propagation scheme would increase the computational complexity and inference latency. Besides, errors can easily propagate in a lengthy video~\cite{zhou2021rta}.

\vspace{0.1in}
\noindent\textbf{Video Deraining}
Video deraining~\cite{jiang2018fastderain, Liu_2018_CVPR, Yang_2019_CVPR_Dual_Flow, RDD, 9439949, Huang_2022_CVPR, patil2022video} is another common low-level vision task that aims to remove rain streaks from videos. Many existing methods used RNN-based approaches to remove diverse spatio-temporal rain patterns~\cite{Liu_2018_CVPR, 9439949, RDD, Yang_2019_CVPR_Dual_Flow}.
Yang~\etal~\cite{9439949} proposed a two-stage recurrent network to recover rain-free images. In the first stage, the network leverages physics guidance to restore rain images. Subsequently, the restored results are further refined through adversarial learning in the second stage.
To exploit the consistency of rain motions in a video, Wang~\etal~\cite{RDD} proposed a recurrent disentanglement-based deraining network to disentangle rain patterns into rain streak maps, background maps, and rain motion maps.

\begin{figure*}
\begin{center}
\includegraphics[width=1\textwidth]{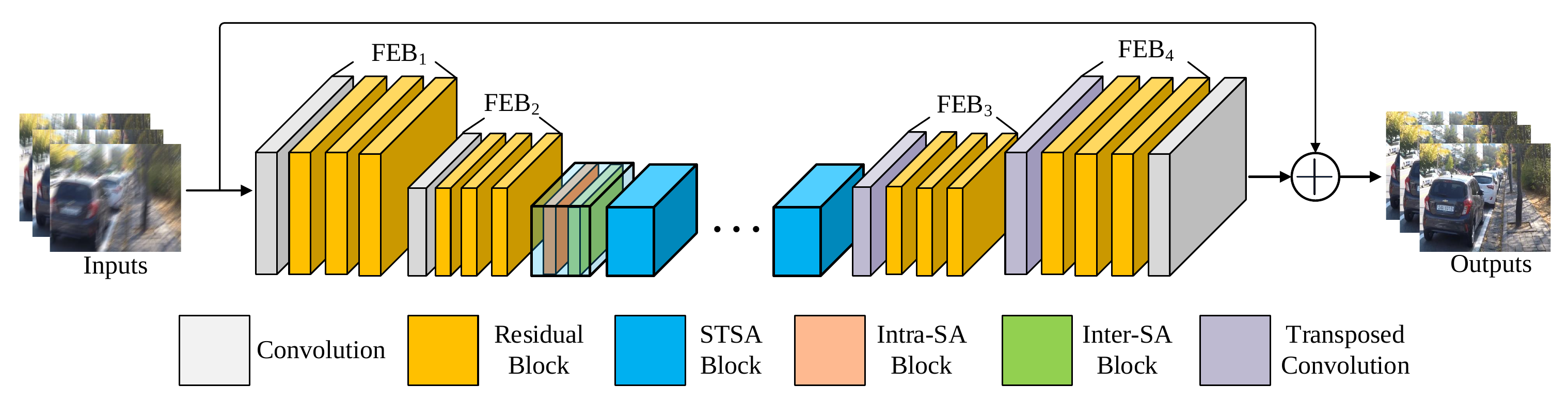}
\end{center}
%\vspace{-0.2in}
\caption{Architecture overview of ViStripformer. We adopt an encoder-decoder structure backboned with Resblocks and fuse spatio-temporal features with STSA blocks, consisting of intra-frame strip attention (Intra-SA) and inter-frame strip attention (Inter-SA) blocks. ViStripformer can restore multiple degraded frames at once, which we illustrate an example with three frames here.}
\label{fig:architecture}
%\vspace{-0.1in}    
\end{figure*} 

\vspace{0.1in}
\noindent\textbf{Video Demoiréing}
Moiré aliasing can cause unpleasant visual experiences and interfere with downstream vision tasks. Hence, it is practical and essential to remove such visual artifacts.
Several deep-learning-based approaches have achieved remarkable performances in single-image demoiréing~\cite{8356681, Zheng_2020_CVPR, he2019mop, hefhde2net, liu2020waveletbased, Yu2022TowardsEA}. However, video demoiréing has received less attention and exploration. 
Dai~\etal~\cite{dai2022video} first collected a hand-held video demoiréing dataset for video demoiréing. They also proposed a video demoiréing network that utilizes a pyramid deformable architecture with temporal consistency regularization based on optical-flow-based warping for demoiréing.

In summary, these video restoration methods resort to recurrent models or optical flows to address temporal alignment for video deblurring, deraining, and demoiréing tasks.  
Nonetheless, recurrent networks can easily accumulate errors due to inaccurate predictions, and optical flows estimated from degraded frames may also be incorrect and would certainly incur additional computational costs.

By contrast, the proposed ViStripfromer is a single-pass end-to-end architecture stacked with STSA blocks to capture long-range spatio-temporal features simultaneously, avoiding recurrent architectures' intrinsic drawbacks and possible inaccurate temporal alignments. ViStripfromer exploits horizontal and vertical strip-wise tokens for video restoration, which are particularly suitable for reassembling features in various directions and magnitudes to handle directional degradation patterns, such as motion blurs, rain streaks, and moiré artifacts.

\subsection{Vision Transformers}
Transformer~\cite{vaswani2017attention} was first proposed for natural language processing (NLP); its self-attention mechanism can effectively build long-range token-wise relations.
Recently, Transformer has also achieved great success in several computer vision tasks, including high-level vision~\cite{dosovitskiy2020vit, liu2021Swin, chu2021Twins, chen2022regionvit, DETR, Ranftl21, zhu2021deformable, xie2021segformer} and low-level vision~\cite{Tsai2022Stripformer, IPT, Zamir2021Restormer, Wang_2022_CVPR, fgst, yang2020learning, yan2020sttn} tasks.
Chen~\etal~\cite{IPT} first proposed a transformer-based model for image restoration. Similar to~\cite{dosovitskiy2020vit}, they utilized a large model and large-scale datasets for training. 
Some methods~\cite{Zamir2021Restormer, Tsai2022Stripformer, Wang_2022_CVPR} used efficient transformer-based models without the need of a large-scale training set for image restoration.
Zamir~\etal~\cite{Zamir2021Restormer} proposed Restormer, which performs self-attention in channels rather than the spatial dimension. Wang~\etal~\cite{Wang_2022_CVPR} proposed Uformer, a U-shape architecture using local window attention proposed in Swin Transformer~\cite{chu2021Twins}.

Recently, Liang~\etal~\cite{liang2022vrt} proposed a video restoration transformer, VRT, which progressively catches spatio-temporal features through sequential window attention and optical-flow-based warping. 
Although the window attention mechanism relieves the high demand of memory consumption, it may not fully explore crucial long-range information in a video sequence.
Therefore, we propose ViStripformer, consisting of intra- and inter-frame strip attentions, to extract long-range information in horizontal and vertical directions.
As mentioned previously, ViStripformer is a token-efficient transformer-based architecture with the space complexity of $\mathcal{O}((H^2+W^2)T)+\mathcal{O}((H+W)T^2)$ for strip-wise attention mechanisms, much lower than the vanilla transformer's $\mathcal{O}((HWT)^2)$.

\section{Proposed Method}
\subsection{Overview}

ViStripformer is a transformer-based model with strip-wise tokens, aiming to restore the visual quality of degraded frames using their consecutive frames. As shown in Fig.~\ref{fig:architecture}, it consists of two feature embedding blocks (FEBs) followed by successive spatio-temporal strip attention (STSA) blocks, composed of intra-frame strip attention (Intra-SA) and inter-frame strip attention (Inter-SA) blocks, and two other FEBs. We will detail these blocks below.

\subsection{Feature Embedding Blocks}
A feature embedding block (FEB) contains one convolutional layer followed by three residual blocks to generate feature embedding. It can better preserve spatial pixel correlations than the patch-embedding strategies~\cite{IPT, dosovitskiy2020vit}. In ViStripformer, these FEBs share same weights and process multiple input frames to encode spatial features with two-dimensional (2D) convolutions.

\begin{figure*}[t!]
\begin{center}
\includegraphics[width=1\textwidth]{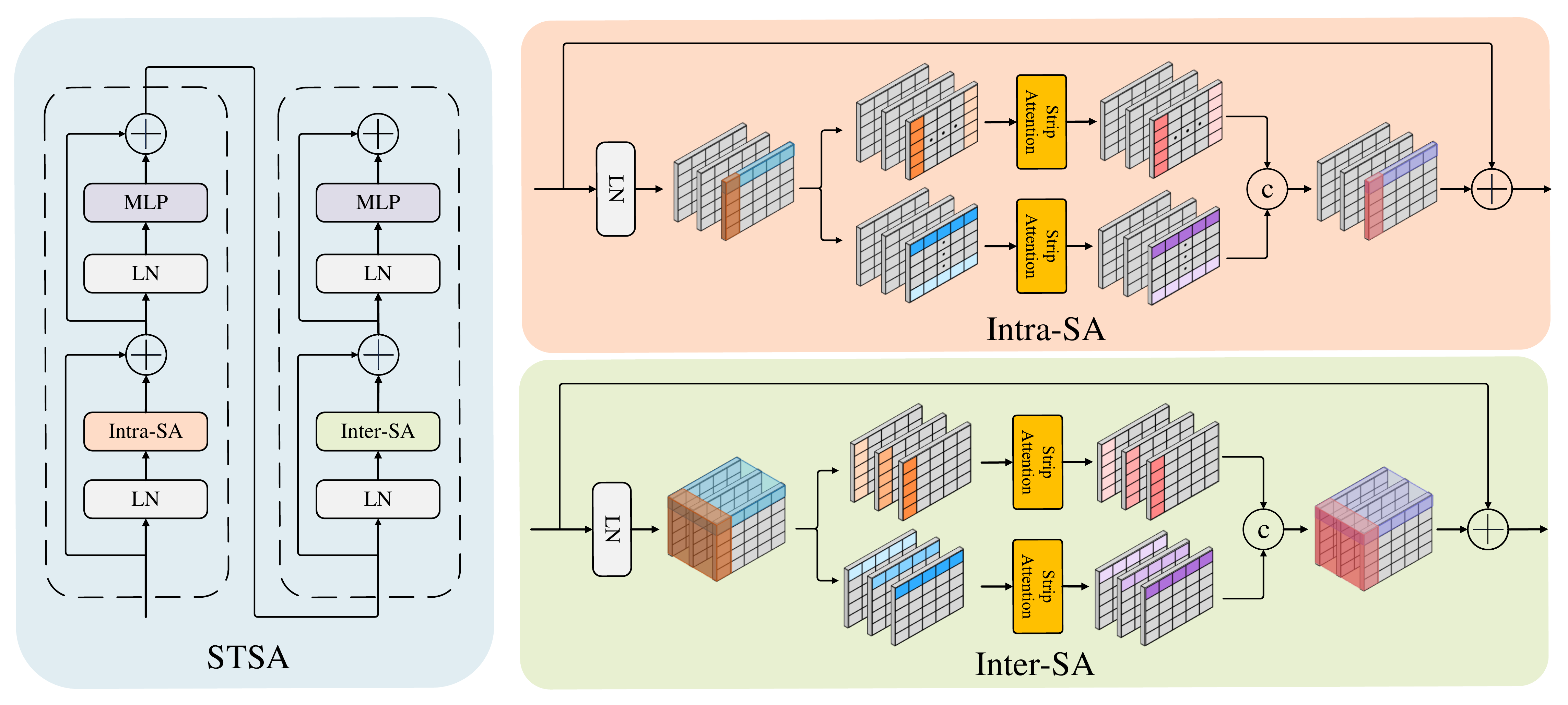}
\end{center}
%\vspace{-0.2in}
\caption{A spatio-temporal strip attention (\textbf{STSA}) block contains an \textbf{Intra-SA} block and an \textbf{Inter-SA} block. Intra-SA performs horizontal and vertical strip-wise attention in a frame. In contrast, Inter-SA performs horizontal and vertical strip-wise attention on each set of the collocated strips across multiple frames, where $\tcircle{c}$ denotes concatenation. 
}
%\vspace{-0.1in}
\label{fig:STSA}
\end{figure*}

\subsection{Spatio-Temporal Strip Attention}
As shown in Fig.~\ref{fig:STSA}, the STSA block cascades an Intra-SA block with an Inter-SA block to extract spatio-temporal features efficiently.

\noindent\textbf{Intra-SA Block}
To efficiently discover long-range spatial dependencies, the proposed Intra-SA block involves two paralleled branches: Horizontal intra-SA (H-Intra-SA) and vertical intra-SA (V-Intra-SA).
Let ${X}$ denote the features of a set of input video frames in $\mathbb{R}^{T\times H \times W \times C}$, where $T$, $H$, $W$, and $C$ are the number of frames, height, width, and the number of channels, respectively.  
For each frame $X_t \in \mathbb{R}^{H \times W \times C},~t\in \{1, 2, ..., T\}$, we first normalize them with Layer Normalization ($\mathrm{LN}$) and split them along channels to acquire input features $X^h_t$ and $X^v_t \in \mathbb{R}^{H \times W \times C^{\prime}}$ for the H-Intra-SA and V-Intra-SA as 
\begin{equation}
    (X^h_t, X^v_t) = \mathrm{Split}(\mathrm{LN}(X_t)), \\
\end{equation}
where $C^{\prime}=\frac{C}{2}$.

In H-Intra-SA, we first generate queries, keys, and values as $Q^{h}_{t,m}$, $K^{h}_{t,m}$, and $V^{h}_{t,m} \in \mathbb{R}^{H \times W \times \frac{C^{\prime}}{M}}$ for the multi-head strip attention as
\begin{equation}
\label{eq:h_i_qkv}
    \begin{gathered} 
    (Q^{h}_{t,m}, K^{h}_{t,m}, V^{h}_{t,m}) = (X^{h}_{t}P^{Q}_{m}, X^{h}_{t}P^{K}_{m},X^{h}_{t}P^{V}_{m}), 
    \end{gathered}
\end{equation}
where $P^{Q}_{m}, P^{K}_{m}$, and $P^{V}_{m} \in \mathbb{R}^{C^{\prime}\times \frac{C^{\prime}}{M}}$, and $m\in \{1, 2, ..., M\}$ respectively denote linear transformation matrices for the queries, keys, and values, and $M=8$ is the number of heads for the following multi-head strip attention.
Next, we reshape $Q^{h}_{t,m}$, $K^{h}_{t,m}$, and $V^{h}_{t,m}$ into 2D tensors in $\mathbb{R}^{H \times \frac{W\times C^{\prime}}{M}}$, representing $H$ horizontal strips in $\mathbb{R}^{\frac{W\times C^{\prime}}{M}}$. The attended features for H-Intra-SA $O^{intra, h}_{t,m} \in \mathbb{R}^{H \times \frac{W\times C^{\prime}}{M}}$ is computed as 
\begin{equation}
    O^{\mathrm{intra, h}}_{t,m} = \mathrm{Softmax}\bigg(\frac{Q^{h}_{t,m}(K^{h}_{t,m})^{\intercal}}{\sqrt{\frac{W\times C^{\prime}}{M}}}\bigg)V^{h}_{t,m},
\end{equation}
and its space complexity is $\mathcal{O}(H^2)$. After that, we reshape $O^{\mathrm{intra, h}}_{t,m}$ back in $\mathbb{R}^{H \times W  \times \frac{C^{\prime}}{M}}$ and then concatenate all the attended features to obtain $O^{\mathrm{intra, h}}_{t}\in \mathbb{R}^{H \times W\times C^{\prime}}$.

Symmetrically, V-Intra-SA computes strip-wise attention in the vertical direction to generate the attended features $O^{\mathrm{intra, v}}_{t,m} \in \mathbb{R}^{W \times \frac{H\times C^{\prime}}{M}}$ with the space complexity of $\mathcal{O}(W^2)$. Likewise, we reshape and concatenate all the V-Intra-SA features as $O^{\mathrm{intra, v}}_{t} \in \mathbb{R}^{H \times W\times C^{\prime}}$.

The final Intra-SA output $O_{t} \in \mathbb{R}^{H \times W\times C}$ is produced as 
\begin{equation}
\label{eq:f_Intra_SA}
\begin{split}
    \tilde{O}^\mathrm{intra}_{t} &= \mathrm{Conv_{1\times1}}(O^{\mathrm{intra, h}}_{t} \oplus O^{\mathrm{intra, v}}_{t}) + X_{t}, \\
    O^\mathrm{intra}_{t} &= \mathrm{MLP}(\mathrm{LN}(\tilde{O}^\mathrm{intra}_{t}) + \tilde{O}^\mathrm{intra}_{t}, \\
\end{split}
\end{equation}
where $\oplus$ denotes concatenation, $\mathrm{Conv_{1\times1}}$ means $1 \times 1$ convolution, and $\mathrm{MLP}$ is a feed-forward Multi-Layer Perceptron.
The space complexity of $\mathrm{Intra}$-$\mathrm{SA}$ for a video frame is $\mathcal{O}(H^2+W^2)$. Thus, the total space complexity for $T$ frames is $\mathcal{O}((H^2+W^2)T)$.

\noindent\textbf{Inter-SA Block.}
For Inter-SA, we apply strip-wise attention to the co-located strips across a window of input frames to extract temporal information. It has two parallel branches: horizontal Inter-SA (H-Inter-SA) and vertical inter-SA (V-Inter-SA). 
Similar to the Intra-SA, we split $X \in \mathbb{R}^{T\times H \times W \times C}$ into $X^h$ and $~X^v$ in $\mathbb{R}^{T \times H \times W \times C^{\prime}}$ after layer normalization to produce the input features for the H-Inter-SA and V-Inter-SA as
\begin{equation}
    (X^h, X^v) = \mathrm{Split}(\mathrm{LN}(X)). \\
\end{equation}

\noindent In H-Inter-SA, $X^h$ is first split into $H$ non-overlapping and co-located horizontal strips $X^h_i \in \mathbb{R}^{T\times W  \times C^{\prime}}, i \in \{1,2, ..., H\}$.
Next, we produce $Q^{h}_{i,m}$, $K^{h}_{i,m}$, and $V^{h}_{i,m} \in \mathbb{R}^{T\times W \times \frac{C^{\prime}}{M}}$ as queries, keys, and values based on a linear projection from $X^h_i$, similar to (\ref{eq:h_i_qkv}).

Next, $Q^{h}_{i,m}$, $K^{h}_{i,m}$, and $V^{h}_{i,m}$ are reshaped into 2D tensors in $\mathbb{R}^{T\times \frac{W \times C^{\prime}}{M}}$. Then,  the strip-wise attention in the temporal dimension generates the attended outputs $O^{\mathrm{inter, h}}_{i,m}\in \mathbb{R}^{T\times \frac{W \times C^{\prime}}{N}}$ as
\begin{equation}
\label{eq:f_Inter_SA}
    O^{\mathrm{inter, h}}_{i,m} = \mathrm{Softmax}\bigg(\frac{Q^{h}_{i,m}(K^{h}_{i,m})^{\intercal}}{\sqrt{\frac{W \times C^{\prime}}{M}}}\bigg)V^{h}_{i,m}.
\end{equation}
with space complexity of $\mathcal{O}(T^2)$.

After that, we reshape $O^{\mathrm{inter, h}}_{i,m}$ into ${T\times W \times \frac{C^{\prime}}{M}}$ for all the heads and concatenate them along the channel dimension to generate $O^{\mathrm{inter, h}}_{i} \in \mathbb{R}^{T\times W  \times C^{\prime}}$. Then, we fold these $H$ attended outputs $O^{\mathrm{inter, h}}_{i}$ into $O^{\mathrm{inter, h}} \in \mathbb{R}^{T\times H\times W\times C^{\prime}}$ as the reweighted features in the horizontal direction.
Symmetrically, we apply the strip-wise attention to $X^v$ for V-Inter-SA and obtain $O^{\mathrm{inter, v}} \in \mathbb{R}^{T\times H\times W\times C^{\prime}}$ as the reweighted features in the vertical direction. The space complexity here is also $\mathcal{O}(T^2)$.

Using the same process to produce the final Intra-SA output in~(\ref{eq:f_Intra_SA}), we then generate the final Inter-SA output $O^{\mathrm{inter}} \in \mathbb{R}^{T\times H\times W\times C}$. Its total space complexity is $\mathcal{O}((H+W)T^2)$ due to simultaneously applying the Inter-SA to the $H$ horizontal strips and $W$ vertical strips. 

With interweaving multi-head Intra-SA and Inter-SA blocks in the proposed ViStripformer, it can better catch degraded patterns with horizontal and vertical strip-shaped features and perform effectively for restoring various video restoration tasks.

\begin{table*}[t!]
\small
\centering
\setlength{\tabcolsep}{3mm}
%\vspace{-0.1in}
\caption{Evaluation results on GoPro test set. The best and second-best scores are \textbf{highlighted} and \underline{underlined}. Params and Time (average per-frame inference time on $1280\times720$ video frames) are measured in (M) and (ms), respectively. $*$ denotes those methods that do not release pre-trained weights or results.}
%\vspace{-0.1in}
\begin{tabular}{l|c|c|c|c|c}
\noalign{\hrule height 1.0pt}
Model & PSNR  & SSIM & Params  & Time & Source   \\
\noalign{\hrule height 1.0pt}
CDVD-TSP~\cite{Pan_2020_CVPR}  & 31.67 & 0.953 & 16 & 1183 & CVPR'20        \\
MSMB~\cite{Ji_2022_CVPR} & 31.77 & 0.949 & \underline{7} & 300  & CVPR'22   \\
STDAN~\cite{zhang2022spatio} & 32.62 & 0.959 & 14 & 990   & ECCV'22  \\
MPRNet~\cite{Zamir2021MPRNet} & 32.66 & 0.959 & 20 & 1410 & CVPR'21   \\
MIMOUNet++~\cite{MIMO} & 32.68 & 0.959 & 16 & 1115 & ICCV'21  \\
MMP-RNN~\cite{wang2022MMP} & 32.75 & 0.960 & \bf{4} & 280 & ECCV'22  \\
FGST$^*$~\cite{fgst} & 32.90 & 0.961  & 10 & 2640 & ICML'22    \\
Restormer~\cite{Zamir2021Restormer} & 32.92 & 0.961 & 26 & 1116 & CVPR'22  \\
Stripformer~\cite{Tsai2022Stripformer} & 33.08 & 0.962 & 20 & 800 & ECCV'22   \\
RNN-MBP~\cite{chao2022} & 33.32 & 0.963 & 16 &  1262 & AAAI'22 \\
ERDN$^*$~\cite{Jiang_2022_ECCV} & 33.48 & 0.933 & 23 & 4900  & ECCV'22    \\
VRT~\cite{liang2022vrt} & 34.81 & 0.972 & 18 & 8625  & arXiv'22\\
RVRT~\cite{liang2022rvrt} & 34.92 & \bf0.974 & 13 & 863  & NIPs'22\\
DSTNet~\cite{Pan_2023_CVPR} & \underline{34.96} & \underline{0.973} & 17 & 278  & CVPR'23 \\
Shfit-Net~\cite{Li_2023_CVPR} & \bf{35.03} & \bf0.974  & \bf4 & 250  & CVPR'23 \\
\hline
ViStripformer & 33.96 & 0.968 & 8 &  \bf{106} & -- \\
ViStripformer+ & 34.93 & \bf{0.974} & 17 & \underline{176}  & --  \\
\noalign{\hrule height 1.0pt}
\end{tabular}
\label{Tab:GoPro}
%\vspace{-0.2in}
\end{table*}

\vspace{0.5em}
\subsection{Loss Function}
The loss functions used in the ViStripformer include a Charbonnier loss and an auxiliary FFT loss. The Charbonnier loss  $\mathcal{L}_\mathrm{Char}$~\cite{Lai_2017_CVPR} is given as
\begin{equation}
\mathcal{L}_\mathrm{Char} = \frac{1}{N}\sum_{i=1}^N\sqrt{||\bf{R}_i - \bf{G}_i||^2 + {\varepsilon}^2},
\end{equation}
where $\bf{R}_i$ and $\bf{G}_i$ respectively denote a restored frame and 
its corresponding ground truth, $N$ is the number of frames restored, and $\varepsilon=10^{-3}$. 
We, in addition, use an auxiliary FFT loss $\mathcal{L}_\mathrm{FFT}$ to reconstruct the high-frequency information for degraded frames as suggested in~\cite{MIMO}. It is written as   
\begin{equation}
\mathcal{L}_\mathrm{FFT} =  \frac{1}{N}\sum_{i=1}^N||\mathcal{F}(\bf{R}_i)-\mathcal{F}(\bf{G}_i)||_1,
\end{equation}
where $\mathcal{F}$ denotes the fast Fourier transform (FFT) function.   
The total loss for the ViStripformer is
\begin{equation}
\mathcal{L}_\mathrm{total} = \mathcal{L}_\mathrm{Char} + \lambda\mathcal{L}_\mathrm{FFT},
\end{equation}
where we set $\lambda=0.01$.

\section{Experiments}

\subsection{Experimental Setups}
We evaluate the proposed ViStripformer\footnote{Our code and additional experimental results can be found in \url{https://github.com/pp00704831/Video-Stripformer}} on three video restoration tasks, including video deblurring, video deraining, and video demoiréing.
We train it using the Adam optimizer with a batch size of $8$. The initial learning rate is $10^{-4}$ and decays to $10^{-7}$ with the cosine annealing strategy. Random cropping, flipping, and rotation are used for data augmentation. 
The number of frames processed at once is 16 for video deblurring and demoiréing and 9 for video deraining.
We implement our model with PyTorch and measure the inference time with the model synchronized time on an NVIDIA 3090 GPU. 
Next, we introduce the datasets used for different tasks in the experiments.

\vspace{0.5em}
\noindent\textbf{Video Deblurring}
We use two benchmark datasets: 1) GoPro~\cite{Nah_2017_CVPR} with $22$ training videos and $11$ testing videos of sharp and synthetic blurred scenes in pairs and 2) BSD (2ms--16ms)~\cite{zhong2022real} with $60$ training videos and $20$ testing videos of real-world blurred and sharp scenes in pairs.

\noindent\textbf{Video Deraining}
Following~\cite{9439949, Huang_2022_CVPR}, we measure the performance on two benchmark datasets: 1) RainSynComplex25~\cite{Liu_2018_CVPR} containing $190$ training videos and $25$ testing videos with heavy rain streaks and 2) RainSynAll100~\cite{9439949} consisting of $900$ training videos and $100$ testing videos with four kinds of degradation: streaks, occlusions, rain accumulation, and accumulation flow.

\noindent\textbf{Video Demoiréing}
Dai~\etal~\cite{dai2022video} collected a video demoiréing dataset, named TCL, using smartphone cameras to capture high-quality videos with various scenes displayed on a screen. The TCL dataset contains $290$ $720\times1280$ videos, each with $60$ frames. We use $247$ training videos and $43$ testing videos following~\cite{dai2022video}.  
%($14,820$ pairs) %($2,580$ pairs)

\begin{table*}[t!]
\small
\centering
\setlength{\tabcolsep}{3mm}
\caption{Evaluation results on BSD (2ms--16ms) test set. The best and the second-best scores are \textbf{highlighted} and \underline{underlined}. The resolution of  BSD  is $640\times480$.}
%\vspace{-0.1in}
\begin{tabular}{l|c|c|c|c|c}
\noalign{\hrule height 1.0pt}
Model & PSNR  & SSIM & Params  & Time  & Source   \\
\noalign{\hrule height 1.0pt}
ESTRNN~\cite{zhong2022real}  & 31.95 & 0.925 & \bf{3} & \bf{26} & IJCV'22        \\
MMP-RNN~\cite{wang2022MMP} & 32.81 & 0.948 & \underline{4} & 61  & ECCV'22    \\
STDAN~\cite{zhang2022spatio} & \underline{33.27} & \underline{0.953} & 14 & 347  & ECCV'22   \\
\hline
ViStripformer & 33.26 & 0.952 & 8 & \underline{36} & --  \\
ViStripformer+ & \bf{33.45} & \bf{0.956} & 17 & 58 & --  \\

\noalign{\hrule height 1.0pt}
\end{tabular}
\label{Tab:BSD}
%\vspace{-0.2in}
\end{table*}

\begin{figure*}[t!]
\begin{center}
\includegraphics[width=2\columnwidth]{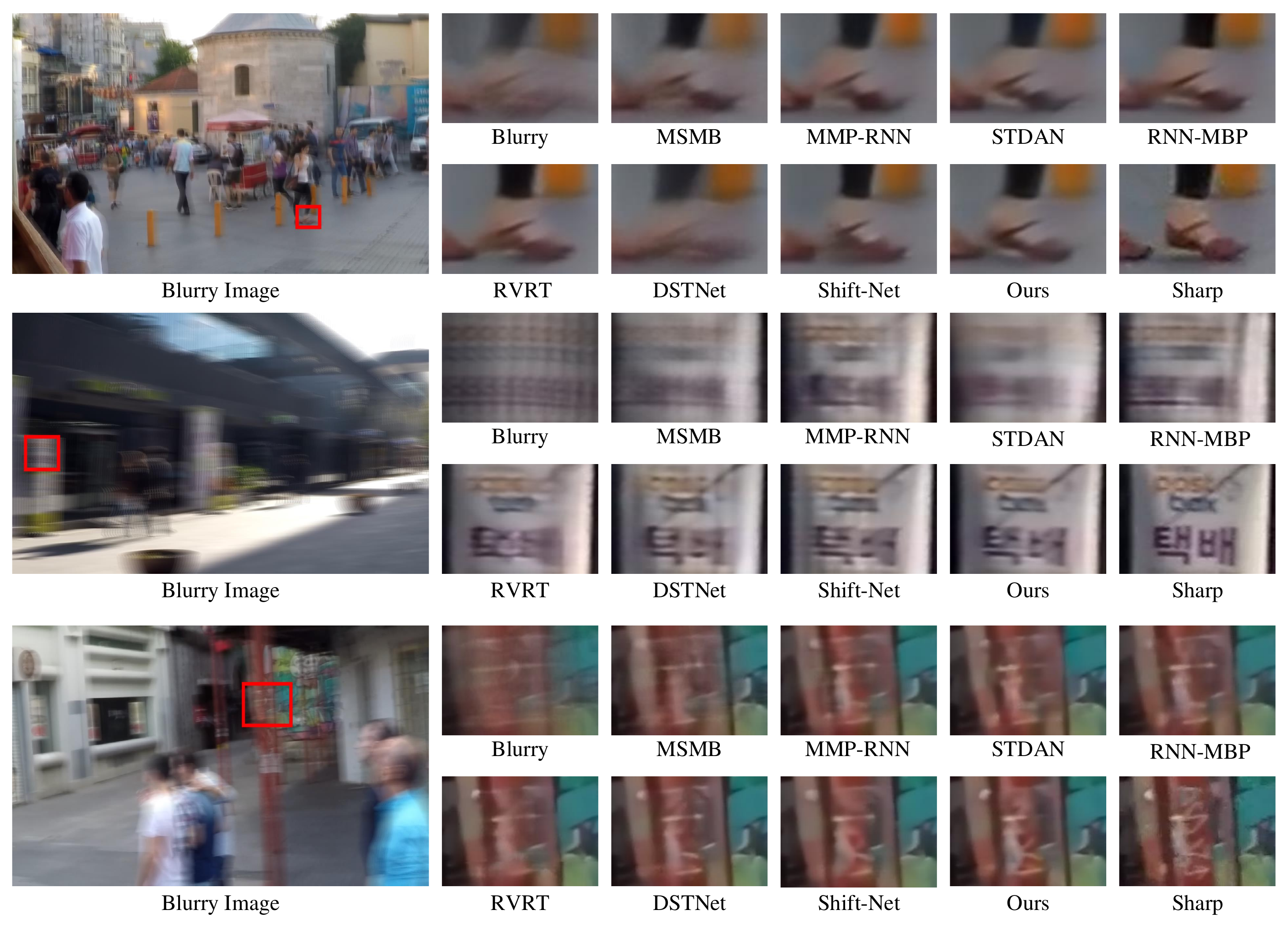}
\end{center}
%\vspace{-0.2in}
\caption{Qualitative comparison of MSMB~\cite{Ji_2022_CVPR}, MMP-RNN~\cite{wang2022MMP}, STDAN~\cite{zhang2022spatio}, RNN-MBP~\cite{chao2022}, RVRT~\cite{liang2022rvrt}, DSTNet~\cite{Pan_2023_CVPR}, Shift-Net~\cite{Li_2023_CVPR}, and Ours (ViStripformer+) on GoPro test set. 
}
\label{fig:gopro}
%\vspace{-0.1in}
\end{figure*}

\subsection{Video Deblurring Results}
We evaluate the proposed method on the synthetic GoPro and real-world BSD (2ms--16ms) test sets for video deblurring. 
Table~\ref{Tab:GoPro} shows the deblurring performance on the GoPro test set.
To better assess the performance of the proposed model for deblurring, we create two versions of our ViStripformer with different model sizes: (1) ViStripformer stacking four STSA blocks (8M parameters), and (2) ViStripformer+ stacking ten STSA blocks (17M). 
As seen in Table~\ref{Tab:GoPro}, ViStripformer+ achieves competitive performance with faster inference time compared to the state-of-the-art VRT~\cite{liang2022vrt}, RVRT~\cite{liang2022rvrt}, DSTNet~\cite{Pan_2023_CVPR}, and Shift-Net~\cite{Li_2023_CVPR}. Note that we also include some recent image-based deblurring methods~\cite{Zamir2021MPRNet, MIMO, Zamir2021Restormer, Tsai2022Stripformer} for comparisons and references.
Compared to the light-weight models, Vistripformer outperforms MSMB~\cite{Ji_2022_CVPR} (7M) and MMP-RNN (4M)~\cite{wang2022MMP}  by 2.17 dB and 1.19 dB in PSNR, respectively, and runs faster.
\begin{table*}[t!]
\centering
\setlength{\tabcolsep}{0.5mm}
\small
\caption{Evaluation results on the RainSynAll100 (ALL100) and  RainSynComplex25 (Complex25) test sets. The best and the second-best scores are \textbf{highlighted} and \underline{underlined}. The other results, except RDD~\cite{RDD} and MVR~\cite{patil2022video}, are provided by NCFL~\cite{Huang_2022_CVPR}. Ours denotes ViStripformer.
}
%\vspace{-0.1in}
\begin{tabular}{lcccccccccc}
\noalign{\hrule height 1.0pt}
 &   & FastDerain~\cite{jiang2018fastderain} & J4RNet~\cite{Liu_2018_CVPR} & FCRVD~\cite{Yang_2019_CVPR_Dual_Flow} & RDD~\cite{RDD} & RMFD~\cite{9439949} & 
 NCFL~\cite{Huang_2022_CVPR} & MVR~\cite{patil2022video}
& Ours \\
\noalign{\hrule height 1.0pt}
\multirow{2}{*}{All100} & PSNR & 17.09 & 19.26 & 21.06 & -- & 25.14 & 28.11 & \underline{28.39} & \bf28.49 \\  
& SSIM & 0.582 & 0.624 & 0.741 & -- & 0.917 & 0.924 & \underline{0.932} & \bf0.943\\ 
\noalign{\hrule height 1.0pt}
%& Time & -- & -- & -- & -- &  & -- &  & \bf57 \\ \hline
\multirow{2}{*}{Complex25} & PSNR & 19.25 & 24.13 & 27.72 & 32.39 & 32.70 & \underline{34.27} & -- & \bf35.09 \\  & SSIM & 0.539 & 0.716 & 0.824 & 0.932 & 0.936 & \underline{0.943} & -- & \bf{0.977}\\ 
%& Time & -- & -- & -- & 1267 & 228 & -- & -- & \bf95 \\
\noalign{\hrule height 1.0pt}
 & Source & TIP'18 & CVPR'18 & CVPR'19 & ECCV'22 & PAMI'22 & CVPR'22 & ECCV'22 & -- \\  
 \noalign{\hrule height 1.0pt}
\end{tabular}
\label{Tab:Deraining}
\vspace{-0.1in}
\end{table*}

\begin{figure*}[t!]
\begin{center}
\includegraphics[width=2\columnwidth]{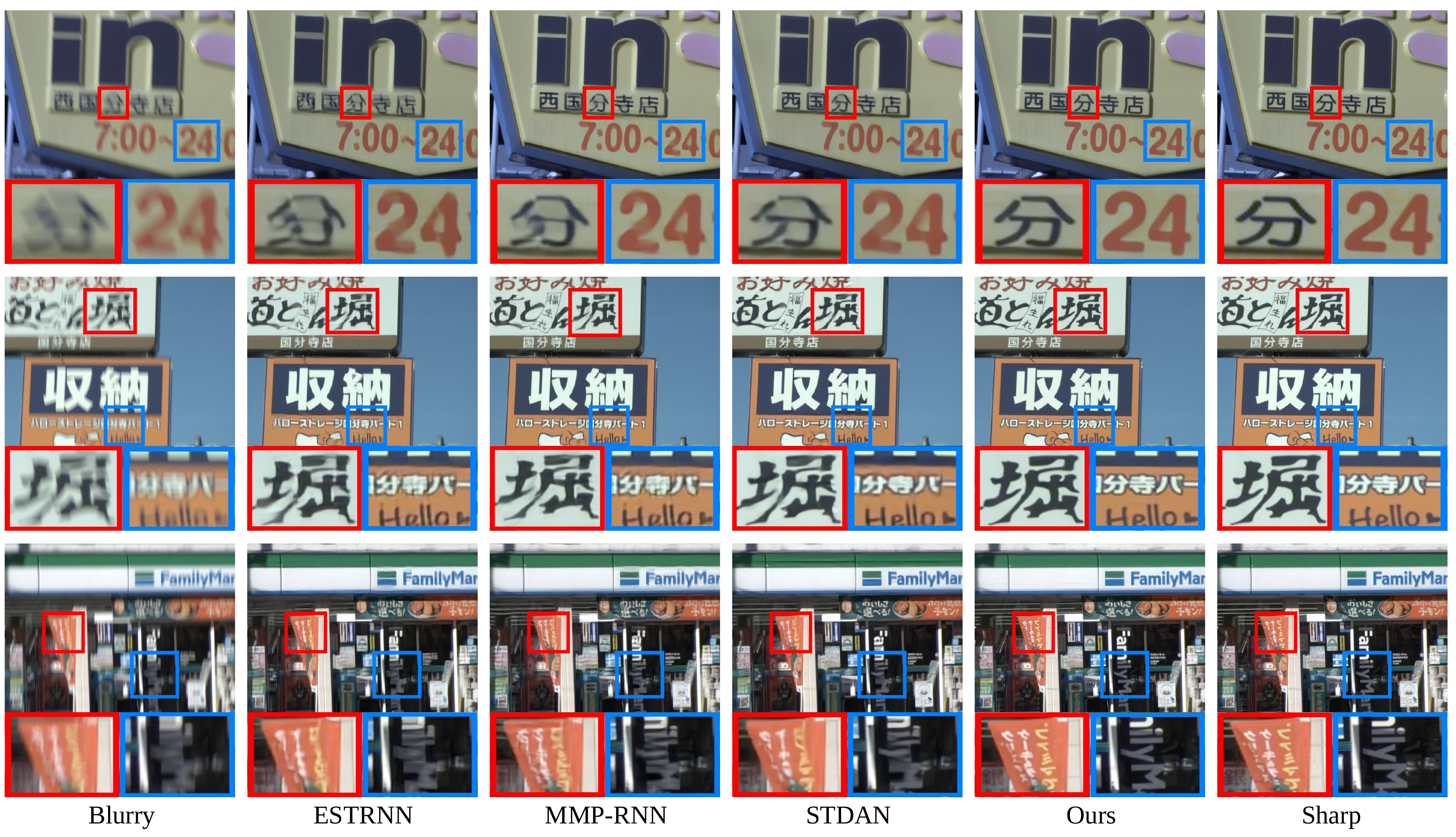}
\end{center}
%\vspace{-0.2in}
\caption{Qualitative comparison of ESTRNN~\cite{zhong2022real}, MMP-RNN~\cite{wang2022MMP}, STDAN~\cite{zhang2022spatio}, and Ours (ViStripformer) on BSD (2ms--16ms) test set.  
}
\label{fig:bsd}
%\vspace{-0.1in}
\end{figure*}

For the real-world BSD (2ms--16ms) test set, we choose three methods that publicly release their results on the BSD (2ms--16ms) testing set, including ESTRNN~\cite{zhong2022real}, MMP-RNN~\cite{wang2022MMP}, and STDAN~\cite{zhang2022spatio}.
ViStripformer outperforms the two RNN models, ESTRNN~\cite{zhong2022real} and MMP-RNN~\cite{wang2022MMP}, shown in Table~\ref{Tab:BSD}. It achieves comparable deblurring performance to STDAN~\cite{zhang2022spatio} with much faster inference time and fewer parameters since STDAN~\cite{zhang2022spatio} uses a recurrent deblurring technique, called cascaded progressive deblurring~\cite{nah2019recurrent}, to improve performance, but it comes with a price of the increased latency.
In Fig.~\ref{fig:gopro}, we qualitatively compare our method with seven video deblurring methods~\cite{Ji_2022_CVPR, zhang2022spatio, wang2022MMP, chao2022, liang2022rvrt, Pan_2023_CVPR, Li_2023_CVPR} on the GoPro test set. Our ViStripformer+ restores blurry images better, as demonstrated in the image patches with the shoe, scene text, and iron pillar.
%
%Besides, since VRT~\cite{liang2022vrt} and RVRT~\cite{liang2022rvrt} have a higher memory demand and space complexity, they require patch cropping during testing, which may cause artifacts at the patch boundaries. In contrast, ViStripformer+ consumes less complexity and thus allows processing full-resolution videos at once without such artifacts. 
A qualitative comparison in Fig.~\ref{fig:bsd} also illustrates that our method can deblur real-world blurred video well compared to the previous methods~\cite{zhong2022real, wang2022MMP, zhang2022spatio} on the BSD (2ms--16ms) test set.
%
%Our method successfully reconstructs the content and details of text, showcasing its ability to handle real-world blurry patterns. All the above experiments show that Video Stripformer successfully addresses video deblurring on both synthetic and real-world blur datasets. 

\begin{table*}[t!]
\centering
\setlength{\tabcolsep}{1mm}
\small
\caption{Evaluation results on the TCL test set. The best and the second-best scores are \textbf{highlighted} and \underline{underlined}. Ours denotes ViStripformer. 
}
%\vspace{-0.1in}
\begin{tabular}{cccccc}
\noalign{\hrule height 1.0pt}
 & MSCNN~\cite{8356681} & MBCNN~\cite{Zheng_2020_CVPR} & VDM~\cite{dai2022video} & ESDNet~\cite{Yu2022TowardsEA}  & Ours \\
\noalign{\hrule height 1.0pt}
PSNR & 20.32 & 21.53 & 21.77 & \underline{22.13} & \bf{22.32}  \\ 
SSIM & 0.703 & 0.740 & 0.729 & \underline{0.780} & \bf{0.795}  \\ 
Source & TIP'18 & CVPR'20 & CVPR'22 & ECCV'22 & --  \\ 
%Time & \bf{62} & 154 & \underline{98} &  106 \\ 
\noalign{\hrule height 1.0pt}
\end{tabular}
\label{Tab:Demoiréing}
%\vspace{-0.15in}
\end{table*}

\begin{figure*}[t!]
\begin{center}
\includegraphics[width=2\columnwidth]{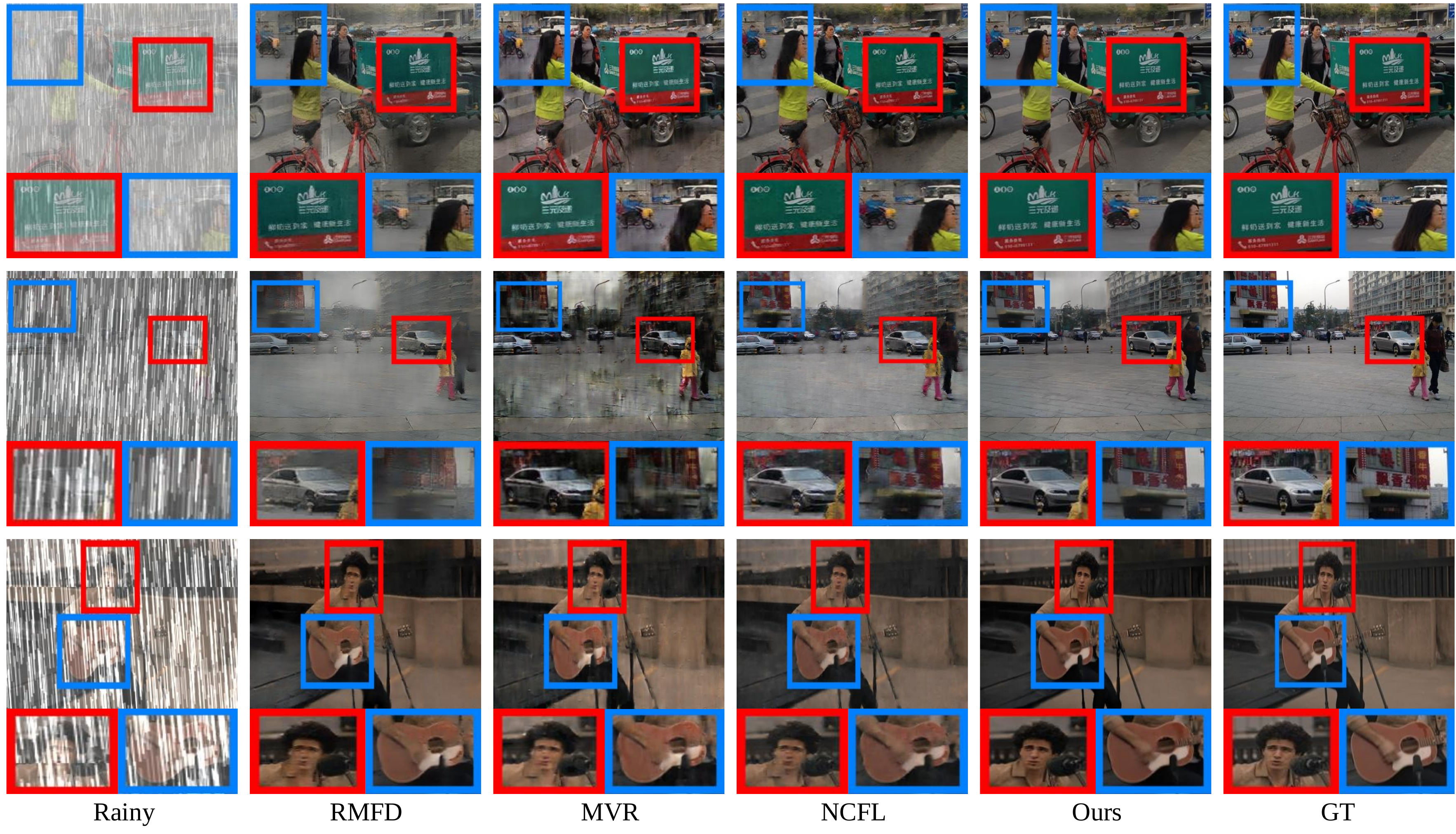}
\end{center}
%\vspace{-0.2in}
\caption{Qualitative comparison of RMFD~\cite{9439949}, MVR~\cite{patil2022video}, NCFL~\cite{Huang_2022_CVPR}, and our method (ViStripformer) on the RainSynAll100 test set. 
}
\label{fig:rain}
%\vspace{-0.1in}
\end{figure*}

\subsection{Video Deraining Results}
Table~\ref{Tab:Deraining} shows quantitative comparisons between ViStripformer and state-of-the-art methods~\cite{jiang2018fastderain, Liu_2018_CVPR, Yang_2019_CVPR_Dual_Flow, RDD, 9439949, Huang_2022_CVPR, patil2022video} on the RainSynAll100~\cite{9439949} and RainSynComplex25~\cite{Liu_2018_CVPR} datasets. 
RainSynAll100 simulates rain steaks based on rain accumulation degradation. The synthetic rain images involve both rain steaks and hazy artifacts, which are more complex than general rain scenes. 
%Thus, some existing methods~\cite{jiang2018fastderain, Liu_2018_CVPR, Yang_2019_CVPR_Dual_Flow} cannot handle these rain videos.
%
%In contrast, RMFD~\cite{9439949}, NCFL~\cite{Huang_2022_CVPR}, MVR~\cite{patil2022video}, and our method can better deal with this degradation. 
%
RainSynComplex25 contains synthetic rain images with heavy rain. 
As observed in Table~\ref{Tab:Deraining}, our method achieves the state-of-the-art results on both the RainSynAll100 and RainSynComplex25 test sets since our ViStripformer utilizes horizontal and vertical strip-shaped tokens in the spatial and temporal dimensions to better remove rain streaks in rain videos.
Fig.~\ref{fig:rain} demonstrates visual comparisons of our method and three past methods, RMFD~\cite{9439949}, MVR~\cite{patil2022video}, and NCFL~\cite{Huang_2022_CVPR}.
The derained results generated by the past methods still present noticeable distortions even though most rain is removed while ViStripformer effectively eliminates rain and haze and yields visually pleasing results.     

\begin{figure*}[t!]
\begin{center}
\includegraphics[width=2\columnwidth]{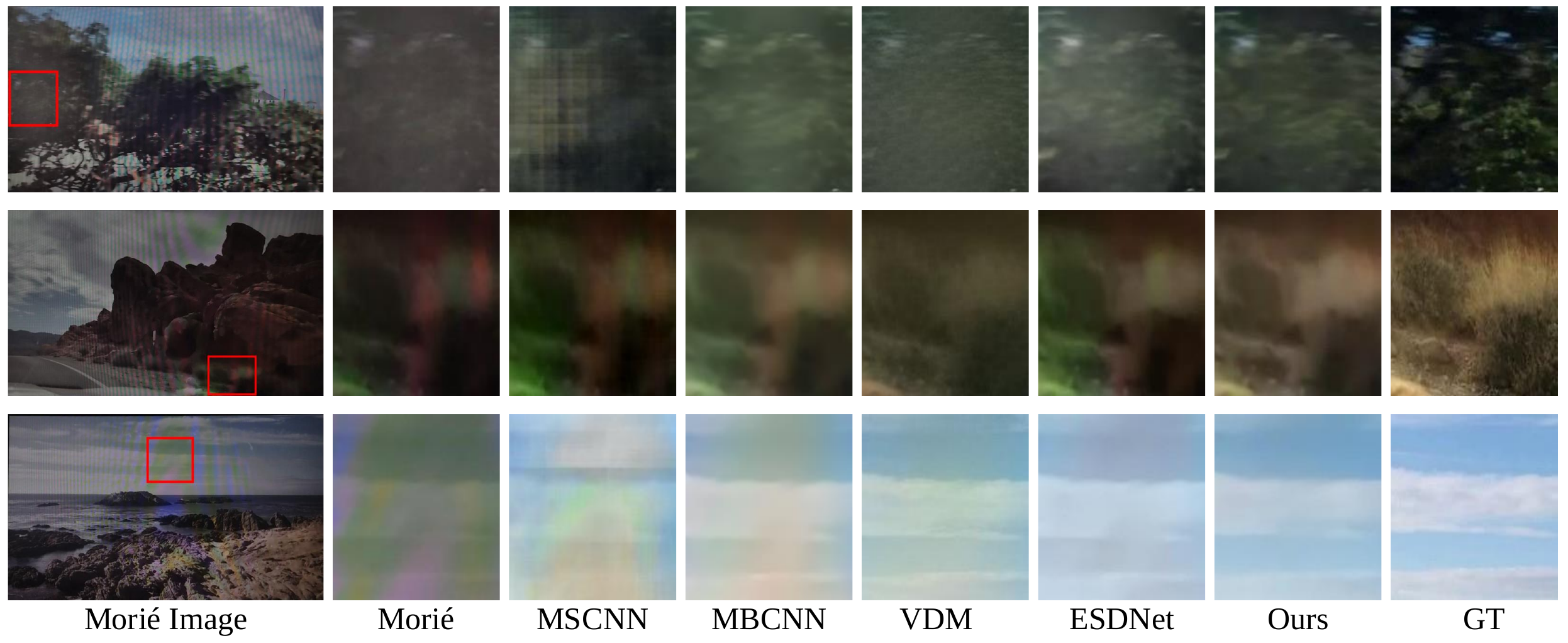}
\end{center}
%\vspace{-0.2in}
\caption{Qualitative comparison of MSCNN~\cite{8356681}, MBCNN~\cite{Zheng_2020_CVPR}, VDM~\cite{dai2022video}, ESDNet~\cite{Yu2022TowardsEA}, and our method (ViStripformer) on the TCL test set.
}
\label{fig:demoering}
%\vspace{-0.1in}
\end{figure*}

\subsection{Video Demoiréing Results}
Table~\ref{Tab:Demoiréing} quantitatively compares the demoiréing performances of our method with four existing methods, including MSCNN~\cite{8356681}, MBCNN~\cite{Zheng_2020_CVPR}, VDM~\cite{dai2022video}, and ESDNet~\cite{Yu2022TowardsEA} on the TCL test set.
The results show that ViStripformer performs favorably against the compared demoiréing methods as it beats the state-of-the-art VDM~\cite{dai2022video} and ESDNet~\cite{Yu2022TowardsEA} by $0.55$ dB and $0.19$ dB in PSNR, respectively.
Fig.~\ref{fig:demoering} presents visual comparisons of MSCNN~\cite{8356681}, MBCNN~\cite{Zheng_2020_CVPR}, VDM~\cite{dai2022video}, ESDNet~\cite{Yu2022TowardsEA}, and ViStripformer.
Our method successfully suppresses Moiré patterns, whereas the other methods still bear unpleasant color distortions caused by Moiré.
In sum, these experimental results clearly demonstrate that the proposed ViStripformer is an effective restoration model to remove directional and strip-shaped artifacts in multiple vision tasks, including video deblurring, deraining, and dmoiréing.

\begin{table}[t]
\centering
\small
\setlength{\tabcolsep}{4mm}
%\vspace{-0.1in}
\caption{Componential analysis of ViStripformer on the GoPro test set. "Joint" refers to performing the strip attention on strip-shape tokens among all frames simultaneously}
\begin{tabular}{ccc|c}
\noalign{\hrule height 1.0pt}
Intra-SA & Inter-SA & Joint & PSNR  \\
\noalign{\hrule height 1.0pt}
$\surd$ &   &   & 32.13 \\
    &  $\surd$  &   & 33.26 \\
    &  &   $\surd$  & 33.76 \\
  $\surd$  &  $\surd$  &    & 33.96 \\
\noalign{\hrule height 1.0pt}
\end{tabular}
\label{tab:ablation}
\end{table}

\begin{table}[t]
\centering
\setlength{\tabcolsep}{4mm}
\caption{Ablation study on the strip-wise attention mechanisms in different directions for ViStripformer on the GoPro test set.}
\vspace{-0.1in}
\begin{tabular}{c|ccc}
\noalign{\hrule height 1.0pt}
 &  Horizontal  & Vertical & Both \\
\noalign{\hrule height 1.0pt}
PSNR & 32.90 &  33.55  & 33.96  \\
\noalign{\hrule height 1.0pt}
\end{tabular}
\label{tab:direction}
\end{table}

\vspace{-0.05in}
\subsection{Ablation Studies}
\begin{figure}[t!]
\begin{center}
\includegraphics[width=1\columnwidth]{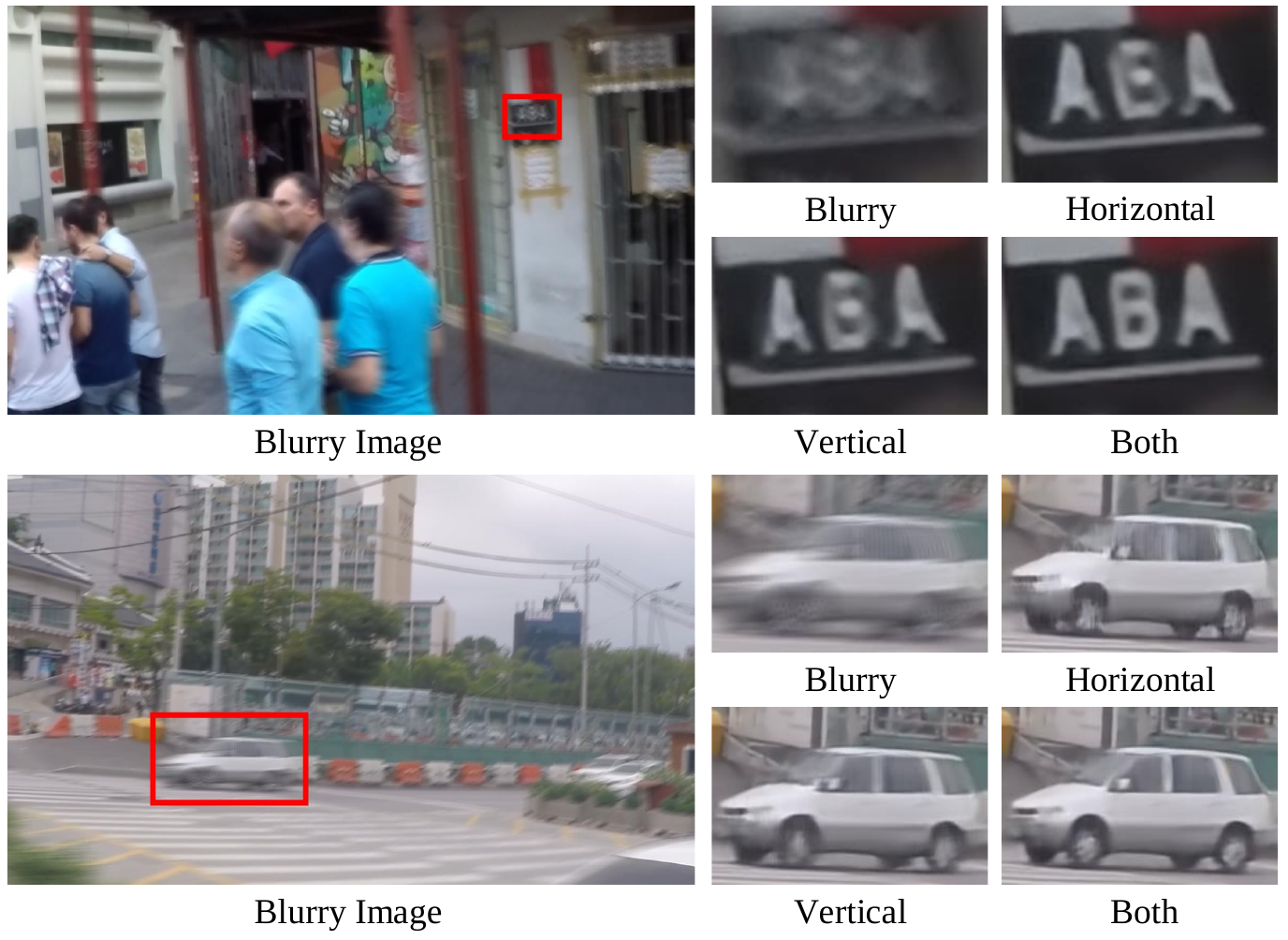}
\end{center}
%\vspace{-0.2in}
\caption{Qualitative comparison of using strip-wise attention mechanisms in different directions for ViStripformer on the GoPro test set. 
}
\label{fig:direction}
\end{figure}

In the ablation studies, we take video deblurring as an example and conduct experiments on the GoPro dataset using ViStripformer with four STSA blocks. As shown in Table~\ref{tab:ablation}, we investigate the effectiveness of the proposed attention mechanism, including Intra-SA and Inter-SA. 
As shown in Fig.~\ref{fig:STSA}, Intra-SA performs strip-wise attention within a frame for $T$ frames, resulting in space complexity of $\mathcal{O}(T(H^2+W^2))$ while Inter-SA operates the attention on each set of the collocated strip-wise tokens across $T$ frames with $\mathcal{O}(T^2(H+W))$. ViStripformer performs Intra-SA and Inter-SA separately and alternatively to perform the strip-wise attention. Additionally, we compare the proposed Intra-SA and Inter-SA with a joint spatio-temporal strip attention, which performs the strip-wise attention on strip-shaped tokens within and across $T$ frames simultaneously, with space complexity of $\mathcal{O}(T^2(H^2+W^2))$. As can be seen in Table~\ref{tab:ablation}, Inter-SA works slightly better than Intra-SA, and combining them achieves the best performance, better than the joint spatio-temporal strip attention.
%
%In contrast, the separate version refers to using an Intra-SA followed by an Inter-SA, which performs better than the joint version, as shown in the third and the last rows.
%
Moreover, Table~\ref{tab:direction} compares the effectiveness of the strip-wise attention in different directions for the ViStripformer. It shows that using both horizontal and vertical strip features in the strip-wise attention leads to higher performance than that of either direction used alone since we need the strip-wise attention in both directions to better handle degradation patterns with various orientations in videos.  Fig.~\ref{fig:direction} also demonstrates the effectiveness of combining both horizontal and vertical strip features to capture directional degradation patterns with various orientations.

We further investigate the effectiveness of the number of STSA blocks stacked. As shown in Table~\ref{tab:Blocks}, we stack four STSA blocks denoted as Stack-4. First, we compare it with wider (Stack-4$^{\dag}$) or deeper (Stack-8) versions with the same parameter size, where we increase the channel size of STSA blocks in Stack-4$^{\dag}$. It shows that stacking more blocks performs better than increasing the channel size. To this end, we stack ten STSA blocks to achieve the best result. 
At last, we conduct an ablation study on $\lambda$ in the loss function. Based on the results shown in Table~\ref{tab:hyperparameter}, we empirically set $\lambda=0.01$.

\begin{table}[t]
\centering
\setlength{\tabcolsep}{0.8mm}
\caption{Ablation study of the stacking number of STSA blocks. Stack-4$^{\dag}$ denotes that we enlarge the channel size of STSA blocks in the  Stack-4 model}
%\vspace{-0.1in}
\begin{tabular}{c|ccccc}
\noalign{\hrule height 1.0pt}
 &  Stack-4  & Stack-4$^{\dag}$ & Stack-8 & Stack-10  & Stack-12 \\
\noalign{\hrule height 1.0pt}
PSNR & 33.96 &  34.13  & 34.58 & \bf34.93 & 34.87  \\
Time (ms) & \bf106  &  143     &  151   &  176  & 198 \\
Params (M) &  \bf8 &  14     &  14   & 17  & 20  \\
\noalign{\hrule height 1.0pt}
\end{tabular}
\label{tab:Blocks}
\end{table}

\begin{table}[t]
\centering
\setlength{\tabcolsep}{2mm}
\caption{Ablation study of the hyperparameter $\lambda$ in the loss funciton.}
\begin{tabular}{c|cccc}
\noalign{\hrule height 1.0pt}
 &  $\lambda=0$  & $\lambda=0.1$ & $\lambda=0.01$ & $\lambda=0.001$ \\
\noalign{\hrule height 1.0pt}
PSNR & 33.87 &  33.89  & {\bf 33.96} & 33.91  \\
\noalign{\hrule height 1.0pt}
\end{tabular}
\label{tab:hyperparameter}
\end{table}

\section{Conclusions}  
We proposed ViStripformer for versatile video restoration, which utilizes token-efficient spatio-temporal strip 
attention (STSA) to capture informative spatio-temporal features.
STSA decomposes the input features into horizontal and vertical strip-shaped tokens to apply intra-frame strip attention and inter-frame strip attention mechanisms. It can efficiently and effectively explore long-range data dependencies for restoring degradation patterns with various orientations and magnitudes.
Extensive experimental results have demonstrated that ViStripformer achieves superior results with fast inference time for three video restoration tasks, deblurring, deraining, and demoiréing.
\section{Data Availability Statement}

All the benchmark datasets used in our experiments including two video deblurring datasets (GoPro~\cite{Nah_2017_CVPR} and BSD (2ms--16ms)~\cite{zhong2022real}), two video draining datasets ( RainSynComplex25~\cite{Liu_2018_CVPR}  and RainSynAll100~\cite{9439949}), and one video demoiréing dataset (TCL~\cite{dai2022video}) are publicly accessible through the individual web links.

Besides, our code and additional experimental results can be found in \url{https://github.com/pp00704831/Video-Stripformer}

\bibliography{sn-bibliography}% common bib file

\end{document}